\documentclass[journal]{IEEEtran}
\usepackage[utf8]{inputenc}

\usepackage{url}
\usepackage{amsmath,bm}
\usepackage{tikz}
\usepackage{graphicx}
\usepackage{verbatim}
\usepackage{microtype}

\newcommand{\unk}{\texttt{\textless unk\textgreater} }

\begin{document}

\title{Automatic Speech Recognition with Very Large Conversational Finnish and
       Estonian Vocabularies}

\author{Seppo~Enarvi, Peter~Smit, Sami~Virpioja,
        and~Mikko~Kurimo,~\IEEEmembership{Senior~Member,~IEEE}%
\thanks{Manuscript received January 31, 2017; revised July 7, 2017; accepted
August 7, 2017.
This work was financially supported by the Academy of Finland under the grant
numbers 251170 and 274075, and by Kone Foundation.}
\thanks{The authors work in the Department of Signal Processing and Acoustics at
Aalto University, Espoo, Finland. (e-mail: firstname.lastname@aalto.fi)}%
\thanks{Digital Object Identifier 10.1109/TASLP.2017.2743344}}

\markboth{IEEE/ACM Transactions on Audio, Speech, and Language
          Processing,~Vol.~25, No.~11, November~2017}%
         {Enarvi \MakeLowercase{\textit{et al.}}: ASR with Very Large
          Conversational Finnish and Estonian Vocabularies}

\maketitle

\IEEEoverridecommandlockouts
\IEEEpubid{
\begin{minipage}{\textwidth}\ \\[12pt]
\centering
2329-9290 \copyright~2017 IEEE. Personal use is permitted, but
republication/redistribution requires IEEE permission.\\
See
\url{http://www.ieee.org/publications_standards/publications/rights/index.html}
for more information.
\end{minipage}
}

\begin{abstract}
Today, the vocabulary size for language models in large vocabulary speech
recognition is typically several hundreds of thousands of words. While this is
already sufficient in some applications, the out-of-vocabulary words are still
limiting the usability in others. In agglutinative languages the vocabulary for
conversational speech should include millions of word forms to cover the
spelling variations due to colloquial pronunciations, in addition to the word
compounding and inflections. Very large vocabularies are also needed, for
example, when the recognition of rare proper names is important.

Previously, very large vocabularies have been efficiently modeled in
conventional n-gram language models either by splitting words into subword units
or by clustering words into classes. While vocabulary size is not as critical
anymore in modern speech recognition systems, training time and memory
consumption become an issue when state-of-the-art neural network language models
are used. In this paper we investigate techniques that address the vocabulary
size issue by reducing the effective vocabulary size and by processing large
vocabularies more efficiently.

The experimental results in conversational Finnish and Estonian speech
recognition indicate that properly defined word classes improve recognition
accuracy. Subword n-gram models are not better on evaluation data than word
n-gram models constructed from a vocabulary that includes all the words in the
training corpus. However, when recurrent neural network (RNN) language models
are used, their ability to utilize long contexts gives a larger gain to
subword-based modeling. Our best results are from RNN language models that are
based on statistical morphs. We show that the suitable size for a subword
vocabulary depends on the language. Using time delay neural network (TDNN)
acoustic models, we were able to achieve new state of the art in Finnish and
Estonian conversational speech recognition, 27.1~\% word error rate in the
Finnish task and 21.9~\% in the Estonian task.
\end{abstract}


\section{Introduction}

\IEEEpubidadjcol

\IEEEPARstart{F}{innish} and Estonian are agglutinative languages, meaning that
words are formed by concatenating smaller linguistic units, and a great deal of
grammatical information is conveyed by inflection. Modeling these inflected
words correctly is important for automatic speech recognition, to produce
understandable transcripts. Recognizing a suffix correctly can also help to
predict the other words in the sentence. By collecting enough training data, we
can get a good coverage of the words in one form or another---perhaps names and
numbers being an exception---but we are far from having enough training data to
find examples of all the inflected word forms.

\IEEEpubidadjcol

Another common feature of Finnish and Estonian is that the orthography is
phonemic. Consequently, the spelling of a word can be altered according to the
pronunciation changes in conversational language. Especially Finnish
conversations are written down preserving the variation that happens in
colloquial pronunciation \cite{Enarvi:2013}. Modeling such languages as a
sequence of complete word forms becomes difficult, as most of the forms are very
rare. In our data sets, most of the word forms appear only once in the training
data.

Agglutination has a far more limited impact on the vocabulary size in English.
Nevertheless, the vocabularies used in English language have grown as larger
corpora are used and computers are able to store larger language models in
memory. Moreover, as speech technology improves, we start to demand better
recognition of e.g. proper names that do not appear in the training data.

Modern automatic speech recognition (ASR) systems can handle vocabularies as
large as millions of words with simple n-gram language models, but a second
recognition pass with a neural network language model (NNLM) is now necessary
for achieving state-of-the-art performance. Vocabulary size is much more
critical in NNLMs, as neural networks take a long time to train, and training
and inference times depend heavily on the vocabulary size. While computational
efficiency is the most important reason for finding alternatives to word-based
modeling, words may not be the best choice of language modeling unit with regard
to model performance either, especially when modeling agglutinative languages.

Subword models have been successfully used in Finnish ASR for more than a decade
\cite{Hirsimaki:2006}. In addition to reducing the complexity of the language
model, subword models bring the benefit that even words that do not occur in the
training data can be predicted. However, subwords have not been used for
modeling \emph{conversational} Finnish or Estonian before. Our earlier attempts
to use subwords for conversational Finnish ASR failed to improve over word
models. In this paper, we show how subword models can be used in the FST-based
Kaldi speech recognition toolkit and obtain the best results to date by
rescoring subword lattices using subword NNLMs, 27.1~\% WER for spontaneous
Finnish conversations, and 21.9 \% WER for spontaneous Estonian conversations.
This is the first published evaluation of subwords in conversational Finnish and
Estonian speech recognition tasks.

Our conclusions are slightly different from those earlier published on standard
Finnish and Estonian tasks, where n-gram models based on statistical morphs have
provided a large improvement to speech recognition accuracy
\cite{Siivola:2003,Hirsimaki:2006,Kurimo:2006}. An important reason is that we
are able to use very large vocabularies (around two million words) in the
word-based n-gram models. Recently it has been noticed that the gap between
subword and word models becomes quite small when such a large word vocabulary is
used \cite{Varjokallio:2016}. In our conversational Finnish and Estonian
experiments, word and subword n-gram models performed quite similarly in terms
of evaluation set word error rate. Our new observation is that neural networks
are especially beneficial for modeling subwords---subword NNLMs are clearly
better than word NNLMs trained using the full vocabulary.

Another approach for very large vocabulary speech recognition
is using word classes in the language models. We evaluate different
algorithms for clustering words into classes. Recent comparisons have shown an
advantage in perplexity for the exchange algorithm over Brown clustering, while
clusterings created from distributed word representations have not worked as
well \cite{Botros:2015,Dehdari:2016,Song:2017}. We present additionally a novel
rule-based algorithm that clusters colloquial Finnish word forms, and also
evaluate word error rate. Surprisingly, class-based n-gram models perform better
than word models in terms of perplexity and speech recognition accuracy in
conversational Finnish and Estonian.

Word classes and subword units are especially attractive in NNLMs, because the
vocabulary size has a great impact on the memory consumption and computational
complexity. The size of the input layer projection matrix and the output layer
weight matrix, as well as the time required to normalize the output
probabilities using softmax, have a linear dependency on the vocabulary size.
The output normalization can also be made more efficient by using one of the
several methods that try to approximate the full softmax, either by modifying
the network structure or the training objective. So far the only comparison of
these approximations for large-vocabulary NNLMs that we are aware of is in
\cite{Chen:2016}. They found hierarchical softmax to perform best in terms of
perplexity with a vocabulary of 800,000 words and a feedforward network.

We compare hierarchical softmax, sampling-based softmax, class-based models, and
subword models in speech recognition on languages that are known for very large
vocabularies. Both data sets contain around two million unique word forms. In
our experiments where the training time was limited to 15 days, class-based
NNLMs clearly exceeded the performance of word-based NNLMs in terms of
perplexity and recognition accuracy. The best results were from subword models.
In the Estonian task, the best subword vocabularies were quite large, and the
best result was from a class-based subword model. We also test two methods for
weighting separate language modeling data sets: weighted sampling, which has
already been introduced in \cite{Schwenk:2005} and update weighting, which is a
novel method.

All the neural network language modeling techniques presented in this paper have
been implemented in the open-source toolkit TheanoLM \cite{Enarvi:2016}, which
we hope to lower the threshold of using neural network language models in speech
recognition research.\footnote{\url{https://github.com/senarvi/theanolm}} We
implemented hierarchical softmax \cite{Goodman:2001}, noise-contrastive
estimation \cite{Gutmann:2010}, and BlackOut \cite{Ji:2016} training criteria,
and a lattice decoder that takes advantage of parallel computation using a GPU.

We use a fairly complex recurrent model consisting of an LSTM layer and a
highway network to obtain state-of-the-art results, and run the experiments on
a high-end GPU. Our experiments show that class and subword models are more
attractive than word models for several reasons. They are efficient
computationally and in terms of memory consumption, and they can achieve better
performance than word models. Usually subword vocabularies include all the
individual letters, meaning that any word that uses the same letters can be 
constructed. Class models are restricted to a certain vocabulary, but the
efficiency is not limited by the vocabulary size, so very large vocabularies can
be used.

To summarize, this is the first time Finnish and Estonian subword models have
outperformed word models in conversational speech recognition, even without
limiting the word vocabulary size. We compare word clustering techniques and
show that class-based models outperform full-vocabulary word models in these
tasks. We also present the first comparison of word, class, and subword NNLMs
trained using different softmax approximations, applied to speech recognition.
Finally, we test a novel method for weighting NNLM training corpora.

\section{Class-Based Language Models}

Finnish and Estonian are highly agglutinative languages, so the number of
different word forms that appear in training corpora is huge. The pronunciation
variation in colloquial Finnish is also written down, making it very difficult
to reliably estimate the probability of the rare words in new contexts. If we
can cluster word forms into classes based on in which contexts they appear, we
can get more reliable estimates for the class n-gram probabilities. In a
class-based language model, the probability of a word within its class is
usually modeled simply as the unigram probability of the word in the training
data \cite{Brown:1992}:

\begin{equation}
\label{eq:class-ngram}
\begin{split}
&P(w_t \mid w_{t-n+1} \ldots w_{t-1}) = \\
&P(c(w_t) \mid c(w_{t-n+1}) \ldots c(w_{t-1})) P(w_t \mid c(w_t)),
\end{split}
\end{equation}

\noindent where $c(w)$ is a function that maps a word to a class. This is also
the model that we use in this article.

\subsection{Statistical Methods for Clustering Words into Classes}

A common cost function for learning the word classes is the perplexity of a
class bigram model, which is equivalent to using the log probability objective:

\begin{equation}
\label{eq:class-bigram}
\mathcal{L} = \sum_t [\log P(c(w_t) \mid c(w_{t-1})) + \log P(w_t \mid c(w_t))]
\end{equation}

Finding the optimal clustering is computationally very challenging. Evaluating
the cost involves summation over all adjacent classes in the training data
\cite{Brown:1992}. The algorithms that have been proposed are suboptimal.
Another approach that can be taken is to use knowledge about the language to
group words that have a similar function.

Brown et al. \cite{Brown:1992} start by assigning each word to a distinct class,
and then merge classes in a greedy fashion. A naive algorithm would evaluate the
objective function for each pair of classes. One iteration of the naive
algorithm would on average run in $\mathcal{O}(N_V^4)$ time, where $N_V$ is the
size of the vocabulary. This involves a lot of redundant computation that can be
eliminated by storing some statistics between iterations, reducing the time
required to run one iteration to $\mathcal{O}(N_V^2)$.

To further reduce the computational complexity, they propose an approximation
where, at any given iteration, only a subset of the vocabulary is considered.
Starting from the most frequent words, $N_C$ words are assigned to distinct
classes. On each iteration, the next word is considered for merging to one of
the classes. The running time of one iteration is $\mathcal{O}(N_C^2)$. The
algorithm stops after $N_V - N_C$ iterations, and results in all the words being
in one of the $N_C$ classes.

The exchange algorithm proposed by Kneser and Ney \cite{Kneser:1993} starts from
some initial clustering that assigns every word to one of $N_C$ classes. The
algorithm iterates through all the words in the vocabulary, and evaluates how
much the objective function would change by moving the word to each class. If
there are moves that would improve the objective function, the word is moved to
the class that provides the largest improvement.

By storing word and class bigram statistics, the evaluation of the objective
function can be done in $\mathcal{O}(N_C)$, and thus one word iterated in
$\mathcal{O}(N_C^2)$ time \cite{Martin:1995}. The number of words that will be
iterated is not limited by a fixed bound. Even though we did not perform the
experiments in such a way that we could get a fair comparison of the training
times, we noticed that our exchange implementation needed less time to converge
than what the Brown clustering needed to finish.\footnote{We are using a
multithreaded exchange implementation and stop the training when the cost stops
decreasing. Our observation that an optimized exchange implementation can be
faster than Brown clustering is in line with an earlier comparison
\cite{Botros:2015}.}

These algorithms perform a lot of computation of statistics and evaluations over
pairs of adjacent classes and words. In practice the running times are better
than the worst case estimates, because all classes and words do not follow each
other. The algorithms can also be parallelized using multiple CPUs, on the
expense of memory requirements. Parallelization using a GPU would be difficult,
because that would involve sparse matrices.

The exchange algorithm is greedy so the order in which the words are iterated
may affect the result. The initialization may also affect whether the
optimization will get stuck in a local optimum, and how fast it will converge.
We use the exchange\footnote{\url{https://github.com/aalto-speech/exchange}}
tool, which by default initializes the classes by sorting the words by frequency
and assigning word $w_i$ to class $i \mod N_C$, where $i$ is the sorted index.
We compare this to initialization from other clustering methods.

\subsection{Clustering Based on Distributed Representation of Words}

Neural networks that process words need to represent them using real-valued
vectors. The networks learn the word embeddings automatically. These
\textit{distributed representations} are interesting on their own, because the
network tends to learn similar representation for semantically similar
words \cite{Mikolov:2013:NAACL}. An interesting alternative to statistical
clustering of words is to cluster words based on their vector representations
using traditional clustering methods.

Distributed word representations can be created quickly using shallow networks,
such as the Continuous Bag-of-Words (CBOW) model \cite{Mikolov:2013:ICLR}. We
use word2vec\footnote{\url{https://code.google.com/archive/p/word2vec/}}
to cluster words by creating word embeddings using the CBOW model and clustering
them using k-means.

\subsection{A Rule-Based Method for Clustering Finnish Words}

Much of the vocabulary in conversational Finnish text is due to colloquial
Finnish pronunciations being written down phonetically. There are often several
ways to write the same word depending on how colloquial the writing style is.
Phonological processes such as reductions (``miksi'' $\rightarrow$ ``miks''
[\textit{why}]) and even word-internal sandhi (``menenpä'' $\rightarrow$
``menempä'' [\textit{I will go}]) are often visible in written form. Intuitively
grouping these different phonetic representations of the same word together
would provide a good clustering. While the extent to which a text is colloquial
does provide some clues for predicting the next word, in many cases these word
forms serve exactly the same function.

This is closely related to normalization of imperfect text, a task which is
common in all areas of language technology. Traditionally text normalization is
based on hand-crafted rules and lookup tables. In the case that annotated data is
available, supervised methods can be used for example to expand abbreviations
\cite{Sproat:2001}. When annotations are not available, candidate expansions for
a non-standard word can be found by comparing its lexical or phonemic form to
standard words \cite{Han:2011}. The correct expansion often depends on the
context. A language model can be incorporated to disambiguate between the
alternative candidates when normalizing text. We are aware of one earlier work
where colloquial Finnish has been translated to standard Finnish using both
rule-based normalization and statistical machine translation
\cite{Listenmaa:2015}.

Two constraints makes our task different from text normalization: A word needs
to be classified in the same way regardless of the context, and a word cannot be
mapped to a word sequence. Our last clustering method, \textit{Rules}, is based
on a set of rules that describe the usual reductions and alternations in
colloquial words. We iterate over a standard Finnish vocabulary and compare the
standard Finnish word with every word in a colloquial Finnish vocabulary. If the
colloquial word appears to be a reduced pronunciation of the standard word,
these words are merged into a class. Because all the words can appear in at most
one class, multiple standard words can be merged into one class, but this is
rare. Thus, there will be only a handful of words in each class. Larger classes
can be created by merging the classes produced by this algorithm using some
other clustering technique.

\section{Subword Language Models}

Subword modeling is another effective technique to reduce vocabulary size. We
use the Morfessor method \cite{Creutz:2002,Creutz:2007:1}, which has been
successfully applied in speech recognition of many agglutinative languages
\cite{Kurimo:2006,Creutz:2007:2}. Morfessor is an unsupervised method that uses
a statistical model to split words into smaller fragments. As these fragments
often resemble the surface forms of morphemes, the smallest information-bearing
units of a language, we will use the term ``morph'' for them.

Morfessor has three components: a model, a cost function, and the training and
decoding algorithm. The model consists of a lexicon and a grammar. The lexicon
contains the properties of the morphs, such as their written forms and
frequencies. The grammar contains information of how the morphs can be combined
into words. The Morfessor cost function is derived from MAP estimation with the
goal of finding the optimal parameters $\theta$ given the observed training data
$D_W$:

\begin{equation}
\begin{aligned}
\theta_{MAP} &= \operatorname*{arg\,max}_{\theta}P(\theta \mid D_W) \\
             &= \operatorname*{arg\,max}_{\theta}P(\theta)P(D_W \mid \theta)
\end{aligned}
\end{equation}

The objective function to be maximized is the logarithm of the product
$P(\theta)P(D_W \mid \theta)$. In a semisupervised setting, it is useful to add
a hyperparameter to control the weight of the data likelihood
\cite{Kohonen:2010}:

\begin{equation}
\mathcal{L}(\theta, D_W) = \log P(\theta) + \alpha \log P(D_W \mid \theta)
\end{equation}

We use the hyperparameter $\alpha$ to control the degree of segmentation in a
heuristic manner. This allows for optimizing the segmentation, either for
optimal perplexity or speech recognition accuracy or to obtain a specific size
lexicon. A greedy search algorithm is used to find the optimal segmentation of
morphs, given the training data. When the best model is found, it is used to
segment the language model training corpus using the Viterbi algorithm.

We apply the Morfessor 2.0
implementation\footnote{\url{https://github.com/aalto-speech/morfessor}} of the
Morfessor Baseline algorithm with the hyperparameter extension
\cite{Virpioja:2013}. In the output segmentation, we prepend and append the
in-word boundaries of the morph surface forms by a ``+'' character. For example
the compound word ``luentokalvoja'' is segmented into ``luento kalvo ja'' and
then transformed to ``luento+ +kalvo+ +ja'' [\textit{lecture+ +slide+ +s}]
before language model training. All four different variants of a subword (e.g.
``kalvo'', ``kalvo+'', ``+kalvo'', and ``+kalvo+'') are treated as separate
tokens in language model training. As high-order n-grams are required to provide
enough context information for subword-based modeling, we use variable-length
n-gram models trained using the VariKN
toolkit\footnote{\url{https://github.com/vsiivola/variKN}} that implements the
Kneser-Ney growing and revised Kneser pruning algorithms \cite{Siivola:2007}.

In the speech recognition framework based on weighted finite-state transducers
(FSTs), we restrict the lexicon FST in such a way that only legal sequences
(meaning that a morph can start with ``+'' if and only if the previous morph
ends with a ``+'') are allowed \cite{Smit:2017}. After decoding the ASR results,
the morphs are joined together to form words for scoring.

\section{Neural Network Language Models}

Recurrent neural networks are known to work well for modeling language, as they
can capture the long-term dependencies neglected by n-gram models
\cite{Mikolov:2010}. Especially the subword-based approach should benefit from
this capability of modeling long contexts. In this article we experiment with
language models that are based on LSTMs and highway networks.
These are layer types that use \textit{sigmoid gates} to control information
flow. The gates are optimized along with the rest of the neural network
parameters, and learn to pass the relevant activations over long distances.

LSTM \cite{Hochreiter:1997} is a recurrent layer type. Each gate can be seen as
an RNN layer with two weight matrices, $W$ and $U$, a bias vector $b$, and
sigmoid activation. The output of a gate at time step $t$ is

\begin{equation}
g(x_t, h_{t-1}) = \sigma(W x_t + U h_{t-1} + b),
\end{equation}

\noindent where $x_t$ is the output vector of the previous layer and $h_{t-1}$
is the LSTM layer state vector from the previous time step. When a signal is
multiplied by the output of a sigmoid gate, the system learns to discard
unimportant elements of the vector depending on the gate's input.

An LSTM layer uses three gates to select what information to pass from the
previous time step to the next time step unmodified, and what information to
modify. The same idea can be used to select what information to pass to the next
layer. Highway networks \cite{Srivastava:2015} use gates to facilitate
information flow across many layers. At its simplest, only one gate is needed.
In the feedforward case, there is only one input, $x_t$, and the gate needs only
one weight matrix, $W_\sigma$. The gate learns to select between the layer's
input and its activation:

\begin{equation}
\begin{aligned}
\label{eq:highway-network}
g(x_t) &= \sigma(W_\sigma x_t + b_\sigma) \\
y_t    &= g(x_t) \odot \tanh(W x_t + b) + (1 - g(x_t)) \odot x_t
\end{aligned}
\end{equation}

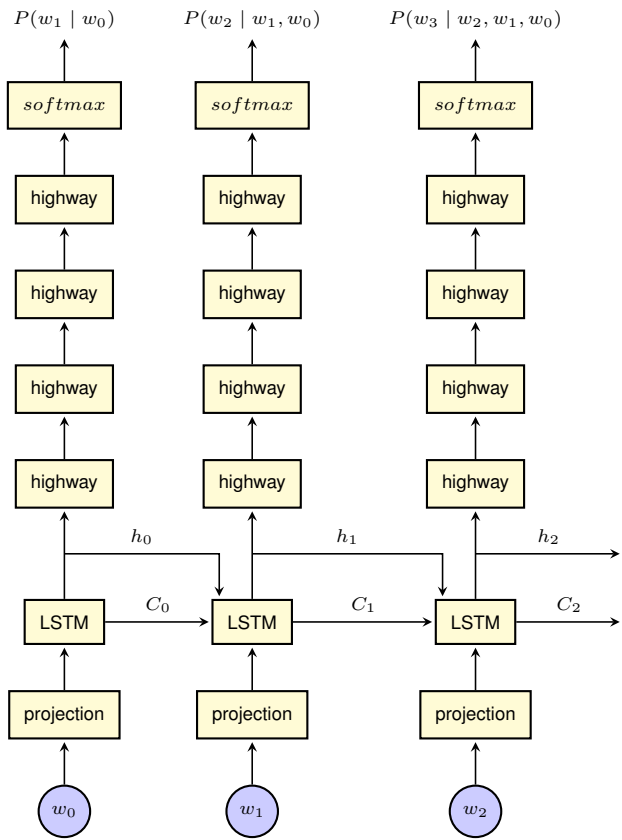
\begin{figure}[t]
\centering
\begin{tikzpicture}[
  font=\sffamily\scriptsize,
  every matrix/.style={ampersand replacement=\&,column sep=7mm,row sep=6mm},
  input/.style={draw,thick,circle,fill=blue!20},
  output/.style={inner sep=.1cm},
  layer/.style={draw,thick,fill=yellow!20,inner sep=.2cm},
  arrow/.style={->,>=stealth,shorten >=1pt,semithick},
  arrow/.style={->,>=stealth,shorten >=1pt,semithick},
  every node/.style={align=center}]

  \matrix {
    \node[output] (y0) {$P(w_1 \mid w_0)$}; \&
    \node[output] (y1) {$P(w_2 \mid w_1, w_0)$}; \&
    \node[output] (y2) {$P(w_3 \mid w_2, w_1, w_0)$}; \&
    \node[coordinate] (y3) {}; \\

    \node[layer] (output0) {$softmax$}; \&
    \node[layer] (output1) {$softmax$}; \&
    \node[layer] (output2) {$softmax$}; \&
    \node[coordinate] (output3) {}; \\

    \node[layer] (hidden40) {highway}; \&
    \node[layer] (hidden41) {highway}; \&
    \node[layer] (hidden42) {highway}; \&
    \node[coordinate] (hidden43) {}; \\

    \node[layer] (hidden30) {highway}; \&
    \node[layer] (hidden31) {highway}; \&
    \node[layer] (hidden32) {highway}; \&
    \node[coordinate] (hidden33) {}; \\

    \node[layer] (hidden20) {highway}; \&
    \node[layer] (hidden21) {highway}; \&
    \node[layer] (hidden22) {highway}; \&
    \node[coordinate] (hidden23) {}; \\

    \node[layer] (hidden10) {highway}; \&
    \node[layer] (hidden11) {highway}; \&
    \node[layer] (hidden12) {highway}; \&
    \node[coordinate] (hidden13) {}; \\

    \node[coordinate] (h0) {}; \&
    \node[coordinate] (h1) {}; \&
    \node[coordinate] (h2) {}; \&
    \node[coordinate] (h3) {}; \\

    \node[layer] (hidden00) {LSTM}; \&
    \node[layer] (hidden01) {LSTM}; \&
    \node[layer] (hidden02) {LSTM}; \&
    \node[coordinate] (hidden03) {}; \\

    \node[layer] (projection0) {projection}; \&
    \node[layer] (projection1) {projection}; \&
    \node[layer] (projection2) {projection}; \&
    \node[coordinate] (projection3) {}; \\

    \node[input] (x0) {$w_0$}; \&
    \node[input] (x1) {$w_1$}; \&
    \node[input] (x2) {$w_2$}; \&
    \node[coordinate] (x3) {}; \\
  };

  \draw[arrow] (x0) -- (projection0);
  \draw[arrow] (x1) -- (projection1);
  \draw[arrow] (x2) -- (projection2);

  \draw[arrow] (projection0) -- (hidden00);
  \draw[arrow] (projection1) -- (hidden01);
  \draw[arrow] (projection2) -- (hidden02);

  \draw[arrow] (hidden00) -- node[midway,above] {$C_0$} (hidden01);
  \draw[arrow] (hidden01) -- node[midway,above] {$C_1$} (hidden02);
  \draw[arrow] (hidden02) -- node[midway,above] {$C_2$} (hidden03);

  \draw[arrow] (h0) -| node[near start,above] {$h_0$} (hidden01.145);
  \draw[arrow] (h1) -| node[near start,above] {$h_1$} (hidden02.145);
  \draw[arrow] (h2) -- node[midway,above] {$h_2$} (h3);

  \draw[arrow] (hidden00) -- (hidden10);
  \draw[arrow] (hidden01) -- (hidden11);
  \draw[arrow] (hidden02) -- (hidden12);

  \draw[arrow] (hidden10) -- (hidden20);
  \draw[arrow] (hidden11) -- (hidden21);
  \draw[arrow] (hidden12) -- (hidden22);

  \draw[arrow] (hidden20) -- (hidden30);
  \draw[arrow] (hidden21) -- (hidden31);
  \draw[arrow] (hidden22) -- (hidden32);

  \draw[arrow] (hidden30) -- (hidden40);
  \draw[arrow] (hidden31) -- (hidden41);
  \draw[arrow] (hidden32) -- (hidden42);

  \draw[arrow] (hidden40) -- (output0);
  \draw[arrow] (hidden41) -- (output1);
  \draw[arrow] (hidden42) -- (output2);

  \draw[arrow] (output0) -- (y0);
  \draw[arrow] (output1) -- (y1);
  \draw[arrow] (output2) -- (y2);

\end{tikzpicture}
\caption{The recurrent network architecture used in our experiments, unrolled
three time steps. The cell state $C_t$ of the LSTM layer conveys information
over time steps until the gates choose to modify it. One gate selects part of
this information as the output of the layer, $h_t$, which is also passed to the
next time step. A highway network uses a gate to select which parts of the
output are passed to the next layer unmodified and which parts are modified.}
\label{fig:network-architecture}
\end{figure}

While LSTM helps propagation of activations and gradients in recurrent networks,
deep networks benefit from highway connections. We did not notice much
improvement by stacking multiple LSTM layers on top of each other. While we did
not have the possibility to systematically explore different network
architectures, one LSTM layer followed by a highway network seemed to perform
well. The architecture used in this article is depicted in Figure
\ref{fig:network-architecture}. Every layer was followed by Dropout
\cite{Srivastava:2014} at rate $0.2$.

The input of the network at time step $t$ is $w_t$, an index that identifies the
vocabulary element. The output contains the predicted probabilities for every
vocabulary element, but only the output corresponding to the target word is
used. The vocabulary can consist of words, word classes, subwords, or subword
classes. The choice of vocabulary does not make any difference with regard to
the neural network training, except that large vocabularies require more memory
and are slower to train.

Usually word vocabularies are limited to a \emph{shortlist} of the most frequent
words. A special token such as \unk can be used in place of any out-of-shortlist
(OOS) words, which is necessary with RNN language models in particular. The NNLM
can be combined with a large-vocabulary n-gram model to obtain a probability for
every training word. The \unk probability represents the total probability mass
of all OOS words, which can be distributed according to n-gram language model
probabilities. An n-gram language model is convenient to integrate with a
feedforward NNLM, which is a particular kind of n-gram model itself, but less
trivial in our RNN decoder. It also becomes computationally demanding to
normalize the resulting probability distribution correctly using a large n-gram
model \cite{Park:2010}.

In our baseline shortlist NNLMs we distribute the OOS probability according to a
unigram model. When rescoring lattices, the output does not have to be a proper
probability distribution. Assuming that the probability mass that the NNLM and
n-gram models allocate to the OOS words are close to each other, a reasonable
approximation is to replace the OOS probabilities with n-gram probabilities
\cite{Park:2010}. We tried this, but did not get good results because the
assumption was too far from the truth. Our class NNLMs are similar to Equation
\ref{eq:class-ngram}, except that RNNs do not fix the context to $n$ previous
words---the length of the history used to predict the next word is limited only
by the mini-batch size.

Memory consumption becomes a problem when using GPUs for training, since current
GPU boards typically have no more than 12 GB of memory. Each layer learns a
weight matrix whose dimensionality is input size by output size. For example, an
NNLM with one hidden layer of size 1,000 and a vocabulary of size 100,000
requires a 1,000 by 100,000 matrix on the input and output layer. Assuming
32-bit floats are used, such a matrix uses 400 MB of memory. In addition,
temporary matrices are needed when propagating the data through the network.
Memory required to store the weight matrices can be reduced by using a small
projection layer and a small layer before the output layer, or factorizing a
large weight into the product of two smaller matrices \cite{Sainath:2013}.
Another possibility is to divide weight matrices to multiple GPUs. The size of
the temporary data depends also on the mini-batch size.

We did the experiments with quite a complex model to see how good speech
recognition accuracy we are able to achieve in these tasks. The projection layer
maps words to 500-dimensional embeddings. Both the LSTM and highway network
layers have 1500 outputs. With vocabularies larger than 100,000 elements we
added a 500-unit feedforward layer before the output layer to reduce memory
consumption. With class models mini-batches were 32 sequences and with other
models 24 sequences of length 25. We also explored the possibility of training
models with larger vocabularies using hierarchical softmax and sampling-based
softmax. These approximations are explained in more detail in the following
sections.

\subsection{Output Normalization}

The final layer normalizes the output to provide a valid probability
distribution over the output classes. Normally the softmax function is used:

\begin{equation}
\label{eq:softmax}
y_{t,i} = \frac{\exp(x_{t,i})}{\sum_j \exp(x_{t,j})}
\end{equation}

At each time step $t$, the cross-entropy cost function requires computing the
conditional probability of the target word only, $P(w_{t+1} \mid w_0 \ldots w_t)
= y_{t,w_{t+1}}$. Still all the activations $x_{t,j}$ are needed to explicitly
normalize the probability distribution. This becomes computationally expensive,
because vocabularies can be very large, and the cost of computing the
normalization term scales linearly with the vocabulary size.

There has been a great deal of research on improving the speed of the softmax
function by various approximations. Hierarchical NNLM is a class-based model
that consists of a neural network that predicts the class and separate neural
networks that predict the word inside a class \cite{Kuo:2012}. This can reduce
training time of feedforward networks considerably, because different n-grams
are used to train each word prediction model. Hierarchical softmax is a single
model that factors the output probabilities into the product of multiple softmax
functions. The idea has originally been used in maximum entropy training
\cite{Goodman:2001}, but exactly the same idea can be applied to neural networks
\cite{Morin:2005}. SOUL combines a shortlist for the most frequent words with
hierarchical softmax for the out-of-shortlist words \cite{Le:2011}. Adaptive
softmax \cite{Grave:2016} is a similar approach that optimizes the word cluster
sizes to minimize computational cost on GPUs.

Another group of methods do not modify the model, but use sampling during
training to approximate the expensive softmax normalization. These methods speed
up training, but use normal softmax during evaluation. Importance sampling is a
Monte Carlo method that samples words from a distribution that should be close
to the network output distribution \cite{Bengio:2003:AISTATS}. Noise-contrastive
estimation (NCE) samples random words, but instead of optimizing the
cross-entropy cost directly, it uses an auxiliary cost that learns to classify a
word as a training word or a noise word \cite{Gutmann:2010}. This allows it to
treat the normalization term as a parameter of the network. BlackOut continues
this line of research, using a stochastic version of softmax that explicitly
discriminates the target word from the noise words \cite{Ji:2016}.

Variance regularization modifies the training objective to encourage the network
to learn an output distribution that is close to a real probability distribution
even without explicit normalization \cite{Shi:2014}. This is useful for example
in one-pass speech recognition, where evaluation speed is important but the
output does not have to be a valid probability distribution. The model can also
be modified to predict the normalization term along with the word probabilities
\cite{Sethy:2015}. NCE objective also encourages the network to learn an
approximately normalized distribution, and can also be used without softmax e.g.
for speech recognition \cite{Chen:2015}.

\subsection{Hierarchical Softmax}
\label{sec:hierarchical-softmax}

Hierarchical softmax factors the output probabilities into the product of
multiple softmax functions. At one extreme, the hierarchy can be a balanced
binary tree that is $log_2(N)$ levels deep, where $N$ is the vocabulary size.
Each level would differentiate between two classes, and in total the
hierarchical softmax would take logarithmic time. \cite{Morin:2005}

We used a two-level hierarchy, because it is simple to implement, and it
does not require a hierarchical clustering of the vocabulary. The first level
performs a softmax between $\sqrt{N}$ word classes and the second level performs
a softmax between $\sqrt{N}$ words inside the correct class:

\begin{equation}
\label{eq:hierarchical-softmax}
\begin{split}
&P(w_t \mid w_0 \ldots w_{t-1}) = \\
&P(c(w_t) \mid w_0 \ldots w_{t-1}) P(w_t \mid w_0 \ldots w_{t-1}, c(w_t))
\end{split}
\end{equation}

\noindent This already reduces the time complexity of the output layer to the
square root of the vocabulary size.

The clustering affects the performance of the resulting model, but it is not
clear what kind of clustering is optimal for this kind of models. In earlier
work, clusterings have been created from word frequencies
\cite{Mikolov:2011:ICASSP}, by clustering distributed word representations
\cite{Mnih:2009}, and using expert knowledge \cite{Morin:2005}.

Ideally all class sizes would be equal, as the matrix product that produces the
preactivations can be computed efficiently on a GPU when the weight matrix is
dense. We use the same word classes in the hierarchical softmax layer that we
use in class-based models, but we force equal class sizes; after running the
clustering algorithm, we sort the vocabulary by class and split it into
partitions of size $\sqrt{N}$. This may split some classes unnecessarily into
two, which is not optimal. On the other hand it is easy to implement and even
as simple methods as frequency binning seem to work \cite{Mikolov:2011:ICASSP}.

An advantage of hierarchical softmax compared to sampling based output layers is
that hierarchical softmax speeds up evaluation as well, while sampling is used
only during training and the output is properly normalized using softmax
during inference.

\subsection{Sampling-Based Approximations of Softmax}

Noise-contrastive estimation \cite{Gutmann:2010} turns the problem from
classification between $N$ words into binary classification. For each training
word, a set of noise words (one in the original paper) is sampled from some
simple distribution. The network learns to discriminate between training words
and noise words. The binary-valued class label $C_w$ is used to indicate whether
the word $w$ is a training or noise word. The authors derive the probability
that an arbitrary word comes from either class, $P(C_w \mid w)$, given the
probability distributions of both classes. The objective function is the
cross entropy of the binary classifier:

\begin{equation}
\begin{aligned}
\mathcal{L} = &\sum_w [C_w \log P(C_w=1 \mid w) \\
              &+ (1 - C_w) \log P(C_w=0 \mid w)]
\end{aligned}
\end{equation}

The expensive softmax normalization can be avoided by making the normalization
term a network parameter that is learned along the weights during training. In a
language model, the parameter would be dependent on the context words, but it
turns out that it can be fixed to a context-independent constant without harming
the performance of the resulting model \cite{Mnih:2012}. In the beginning of the
training the cost will be high and the optimization may be unstable, unless the
normalization is close to correct. We use one as the normalization constant and
initialize the output layer bias to the logarithmic unigram distribution, so
that in the beginning the network corresponds to the maximum likelihood unigram
distribution.

BlackOut \cite{Ji:2016} is also based on sampling a set of noise words, and
motivated by the discriminative loss of NCE, but the objective function
directly discriminates between the training word $w_T$ and noise words $w_N$:

\begin{equation}
\mathcal{L} = \sum_{w_T} [\log P(w_T) + \sum_{w_N} \log (1 - P(w_N))]
\end{equation}

\noindent Although not explicitly shown, the probabilities $P(w)$ are
conditioned on the network state. They are computed using a weighted softmax
that is normalized only on the set of training and noise words. In addition to
reducing the computation, this effectively performs regularization in the output
layer similarly to how the Dropout \cite{Srivastava:2014} technique works in the
hidden layers.

Often the noise words are sampled from the uniform distribution, or from the
unigram distribution of the words in the training data \cite{Mnih:2012}. Our
experiments confirmed that the choice of \textit{proposal distribution} is
indeed important. Using uniform distribution, the neural network optimization
will not find as good parameters. With unigram distribution the problem is that
some words may be sampled very rarely. Mikolov et al. \cite{Mikolov:2013:NIPS}
use the unigram distribution raised to the power of $\beta$. Ji et al.
\cite{Ji:2016} make $\beta$ a tunable parameter. They also exclude the correct
target words from the noise distribution.

We used the power distribution with $\beta = 0.5$ for both BlackOut and NCE.
We did not modify the distribution based on the target words, however, as that
would introduce additional memory transfers by the Theano computation library
used by TheanoLM. We observed also that random sampling from a multinomial
distribution in Theano does not work as efficiently as possible with a GPU. We
used 500 noise words, shared across the mini-batch. These values were selected
after noting the speed of convergence with a few values. Small $\beta$ values
flatten the distribution too much and the optimal model is not reached. Higher
values approach the unigram distribution, causing the network to not learn
enough about the rare words. Using more noise words makes mini-batch updates
slower, while using only 100 noise words we noticed that the training was barely
converging.

These methods seem to suffer from some disadvantages. Properly optimizing the
$\beta$ parameter can take a considerable amount of time. A large enough set of
noise words has to be drawn for the training to be stable, diminishing the speed
advantage in our GPU implementation. While we did try a number of different
parameter combinations, BlackOut never finished training on these data sets
without numerical errors.

\subsection{Decoding Lattices with RNN Language Models}

While improving training speed is the motivation behind the various softmax
approximations, inference is also slow on large networks. Methods that modify
the network structure, such as hierarchical softmax, improve inference speed as
well. Nevertheless, using an RNN language model in the first pass of
large-vocabulary speech recognition is unrealistic. It is possible to create a
list of $n$ best hypothesis, or a word lattice, during the first pass, and
rescore them using an NNLM in a second pass. We have implemented a word lattice
decoder in TheanoLM that produces better results than rescoring n-best lists.

Conceptually, the decoder propagates tokens through the lattice. Each token
stores a network state and the probability of the partial path. At first one
token is created at the start node with the initial network state. The algorithm
iterates by propagating tokens to the outgoing links of a node, creating new
copies of the tokens for each link. Evaluating a single word probability at a
time would be inefficient, so the decoder combines the state from all the tokens
in the node into a matrix, and the input words into another matrix. Then the
network is used to simultaneously compute the probability of the target word in
all of these contexts.

Rescoring a word lattice using an RNN language model is equivalent to rescoring
a huge n-best list, unless some approximation is used to limit the dependency of
a probability on the earlier context. We apply three types of pruning, before
propagation, to the tokens in the node \cite{Sundermeyer:2014:LatticeDecoding}:

\begin{itemize}
  \item \textbf{n-gram recombination}. \enspace If there are multiple tokens,
  whose last $n$ context words match, keep only the best. We use $n = 22$.
  \item \textbf{cardinality pruning}. \enspace Keep at most $c$ best tokens.
  We use $c = 62$.
  \item \textbf{beam pruning}. \enspace Prune tokens whose probability is low,
  compared to the best token. The best token is searched from all nodes that
  appear at the same time instance, or in the future. (Tokens in the past have a
  higher probability because they correspond to a shorter time period.) We prune
  tokens if the difference in log probability is larger than 650.
\end{itemize}

We performed a few tests with different pruning parameters and chose large
enough $n$ and $c$ so that their effect in the results was negligible. Using a
larger beam would have improved the results, but the gain would have been small
compared to the increase in decoding time.

\section{Experiments}

\subsection{Data Sets}

We evaluate the methods on difficult spontaneous Finnish and Estonian
conversations. The data sets were created in a similar manner for both
languages. For training acoustic models we combined spontaneous speech corpora
with other less spontaneous language that benefits acoustic modeling.
For training language models we combined transcribed conversations with web data
that has been filtered to match the conversational speaking style
\cite{Kurimo:2016}.

For the Finnish acoustic models we used 85 hours of training data from
three sources. The first is the complete Finnish SPEECON \cite{Iskra:2002}
corpus. This corpus includes 550 speakers in different noise conditions that all
have read 30 sentences and 30 words, numbers, or dates, and spoken 10
spontaneous sentences. Two smaller data sets of better matching spontaneous
conversations were used: DSPCON \cite{Aalto:2017} corpus, which consists of
short conversations between Aalto University students, and FinDialogue part of
the FinINTAS \cite{Lennes:2009} corpus, which contains longer spontaneous
conversations. For language modeling we used 61,000 words from DSPCON and 76
million words of web data. We did not differentiate between upper and lower
case. This resulted in 2.4 million unique words.

For the Estonian acoustic models we used 164 hours of training data,
including 142 hours of broadcast conversations, news, and lectures collected at
Tallinn University of Technology \cite{Meister:2012}, and 23 hours of
spontaneous conversations collected at the University of Tartu\footnote{Phonetic
Corpus of Estonian Spontaneous Speech. For information on distribution, see
\url{http://www.keel.ut.ee/et/foneetikakorpus}.}. These transcripts contain 1.3
million words. For language modeling we used additionally 82 million words
of web data. The language model training data contained 1.8 million unique
words, differentiating between upper and lower case. One reason why the Estonian
vocabulary is smaller than the Finnish vocabulary, even though the Estonian data
set is larger, is that colloquial Estonian is written in a more systematic way.
Also standard Estonian vocabulary is smaller than standard Finnish vocabulary
\cite{Creutz:2007:2}, probably because standard Finnish uses more inflected word
forms.

\begin{table}[t]
\caption{{\it Out-of-vocabulary word rates (\%) of the evaluation sets,
excluding start and end of sentence tokens. The last row is the full training
set vocabulary, which applies also for the class models.}}
\label{tab:oov-rates}
\begin{center}
\begin{tabular}{lcc}
\hline
Vocabulary Size & Finnish & Estonian \\
\hline
100,000                 & 6.67    & 3.89 \\
500,000                 & 3.36    & 1.59 \\
2.4M (Fin) / 1.8M (Est) & 2.31    & 1.01 \\
\hline
\end{tabular}
\end{center}
\end{table}

We use only spontaneous conversations as development and evaluation data.
As mentioned earlier, Finnish words can be written down in as many different
ways as they can be pronounced in colloquial speech. When calculating Finnish
word error rates we accept the different forms of the same word as
correct, as long as they could be used in the particular context. Compound words
are accepted even if they are written as separate words. However, we compute
perplexities on transcripts that contain the phonetically verbatim word forms,
excluding out-of-vocabulary (OOV) words. The perplexities from n-gram and neural
network word and class models are all comparable to one another, because they
model the same vocabulary consisting of all the training set words. Subwords can
model also unseen words, so the perplexities in subword experiments are higher.
OOV word rates of the evaluation sets are reported in Table \ref{tab:oov-rates}
for different vocabulary sizes.

The Estonian web data is the filtered data from \cite{Kurimo:2016}. The same
transcribed data is also used, except that we removed from the acoustic
training set three speakers that appear in the evaluation set. The evaluation
data is still 1236 sentences or 2.9 hours. The Finnish data is what we used in
\cite{Enarvi:2016}, augmented with 2016 data of DSPCON and read speech from
SPEECON. While we now have more than doubled the amount of acoustic training
data, we have only a few more hours of spontaneous conversations. The switch to
neural network acoustic models had a far greater impact on the results than the
additional training data. We still use the same Finnish evaluation set of 541
sentences or 44 minutes. The Finnish development and evaluation sets and
reference transcripts that contain the alternative forms are included in the
latest DSPCON release, without a few sentences that we could not license.

\subsection{Models}

The word based n-gram models were 4-grams, trained using the Modified Kneser-Ney
implementation of SRILM toolkit \cite{Stolcke:2002}. Class-based models did not
use Kneser-Ney smoothing, because the class n-gram statistics were not suitable
for computing the Modified Kneser-Ney discount parameters. The quality of our
web data is very different from the transcribed conversations, and simply
pooling all the training data together would cause the larger web data to
dominate the model. Instead we created separate models from different data sets,
and combined them by interpolating the probabilities of the observed n-grams
from the component models using weights that were optimized on the development
data. In the Finnish task we created a mixture from two models, a web data model
and a transcribed data model. In the Estonian task we created a mixture from
three models, separating the transcribed spontaneous conversations from the
broadcast conversations.

The mixture weights were optimized independently for each language model on the
development data, using expectation maximization (EM). In the Finnish
experiments this gave the transcribed data a weight slightly less than 0.5. In
the Estonian experiments the weights of the spontaneous conversations and the
web data were typically around 0.4, while the broadcasts were given a weight
less than 0.2. Morph models were similarly combined from component models, but
the EM optimization failed to give good weights. We used initially those
optimized for the word-based models, and after the other parameters were fixed,
we optimized the mixture weights for development set perplexity using a grid
search with steps of $0.05$.

The word clustering algorithms do not support training data weighting, so we
simply concatenated the data sets. There are many parameters that can be
tweaked when creating distributed word representations with word2vec. We tried
clustering words using a few different parameters, and report only the best
n-gram model for each class vocabulary size. Within the set of values that we
tried, the best performance was obtained with continuous bag of words (CBOW),
window size 8, and layer size 300 to 500.

For the subword language models, we trained Morfessor on a word list combined
from all training corpora; the difference to other options such as token-based
training was negligible. For each language, four segmentations were trained with
$\alpha$-values 0.05, 0.2, 0.5, and 1.0. This resulted in respective vocabulary
sizes of 42.5k, 133k, 265k, and 468k for Finnish, and 33.2k, 103k, 212k, and
403k for Estonian. The sizes include the different morph variants with ``+''
prefix and affix. When training the subword n-gram models with the VariKN
toolkit, the growing threshold was optimized on the development set, while
keeping the pruning threshold twice as large as the growing threshold.

Word-based neural network models were trained on two shortlist sizes: 100k and
500k words. With 500k words we added a normal 500-unit layer with hyperbolic
tangent activation before the output layer, which reduced memory consumption and
speeded up training. The neural networks were trained using Adagrad
\cite{Duchi:2011} optimizer until convergence or until the maximum time limit of
15 days was reached. All neural network models were trained on a single NVIDIA
Tesla K80 GPU and the training times were recorded.

We tried two different approaches for weighting the different data sets during
neural network training: by randomly sampling a subset of every data set in the
beginning of each epoch \cite{Schwenk:2005}, and by weighting the parameter
updates depending on from which corpus each sentence comes from. In the latter
approach, the gradient is scaled by both the learning rate and a constant that
is larger for higher-quality data, before updating the parameters. We are not
aware that this kind of update weighting would have been used before.

Optimizing the weights for neural network training is more difficult than
for the n-gram mixture models. As we do not have a computational method for
optimizing the weights, we tried a few values, observing the development set
perplexity during training. Sampling 20 \% of the web data on each iteration, or
weighting the web data by a factor of 0.4 seemed to work reasonably well. We
used a slightly higher learning rate when weighting the web data to compensate
for the fact that the updates are smaller on average. More systematic tests were
performed using these weights with the five vocabularies in Table
\ref{tab:subset-processing-results}.

It would be possible to train separate neural network models from each data set,
but there are no methods for merging several neural networks in the similar
fashion that we combine the n-gram models. Often the best possible NNLM results
are obtained by interpolating probabilities from multiple models, but that kind
of system is cumbersome in practice, requiring multiple models to be trained and
used for inference. The layer sizes and other parameters would have to be
optimized for each model separately.

We combined the NNLMs with the nonclass word or subword n-gram model by
log-linear interpolation. We did not notice much difference to linear
interpolation, so we chose to do the interpolation in logarithmic space, because
the word probabilities may be smaller than what can be represented using 64-bit
floats. We noticed that optimization of the interpolation parameters was quite
difficult with our development data, so we gave equal weight to both models. In
some cases it could have been beneficial to give a larger weight to the neural
network model. Development data was used to select a weight for combining
language model and acoustic scores from four different values.

\subsection{Speech Recognition System}

We use the Kaldi \cite{Povey:2011} speech recognition system for training our
acoustic models and for first-pass decoding. The TDNN acoustic models were
trained on a pure sequence criterion using Maximum Mutual Information (MMI)
\cite{Povey:2016}. The data sets were cleaned and filtered using a Gaussian
Mixture Model recognizer and augmented through speed and volume perturbation
\cite{Ko:2015}. The number of layers and parameters of the TDNN were optimized
to maximize development set accuracy on the word model.
First-pass decoding was very fast with real-time factor less than $0.5$. The
accuracy of the first-pass recognition exceeded our earlier results on both data
sets \cite{Enarvi:2016,Kurimo:2016}, due to the new neural network acoustic
models.

Kaldi does not, at this moment, directly support class-based decoding. Instead
we created lattices using regular n-gram models, and rescored them with class
n-gram and neural network models. Using the methods described in
\cite{Allauzen:2003} it is possible to construct a word FST that represents the
probabilities of a class-based language model and use this in first-pass
decoding or rescoring, without explicitly using the classes. Kaldi does not have
any restrictions on vocabulary size, but compiling the FST-based decoding graph
from the language model, pronunciation dictionary, context dependency
information, and HMM structure did consume around 60 GB of memory. The memory
requirement can be lowered by reducing the number of contexts in the first-pass
n-gram model.

\subsection{Results}

\begin{table}[t]
\caption{{\it Results from word, class, and subword n-gram language models on
development data. Includes perplexity, word error rate (\%), and word error
rate after interpolation with the nonclass model. Exchange algorithm is
initialized using classes created with the other algorithms, in addition to the
default initialization. Perplexities from word and subword models are not
comparable.}}
\label{tab:ngram-dev-results}
\begin{center}
\begin{tabular}{lccc}
\hline
Classes            & Perplexity &    WER & +Nonclass \\
\hline
\noalign{\vskip 1mm}
\multicolumn{4}{l}{\textbf{Finnish, 2.4M words}} \\
Nonclass           &      736 &     30.5 & \\
Brown 2k           &      705 &     29.9 &     29.0 \\
CBOW 2k            &     1017 &     33.2 &     30.2 \\
Exchange 2k        &      698 &     29.7 &     29.1 \\
Brown+Exchange 2k  &      701 &     30.0 &     29.0 \\
CBOW+Exchange 2k   &      695 &     29.8 &     29.3 \\
Rules+Exchange 2k  &      700 & \bf 29.3 & \bf 28.9 \\
Brown 5k           &      694 &     29.9 &     29.3 \\
CBOW 5k            &      861 &     31.4 &     29.8 \\
Exchange 5k        &  \bf 683 &     29.9 &     29.3 \\
Brown+Exchange 5k  &      688 &     29.8 &     29.2 \\
CBOW+Exchange 5k   &      684 &     29.8 &     29.3 \\
Rules+Exchange 5k  &      688 &     29.8 &     29.4 \\
CBOW 10k           &      801 &     31.1 &     29.9 \\
Exchange 10k       &      691 &     29.9 &     29.4 \\
CBOW+Exchange 10k  &      691 &     29.9 &     29.5 \\
Rules+Exchange 10k &      690 &     29.9 &     29.4 \\[1mm]
\multicolumn{4}{l}{\textbf{Finnish, 42.5k subwords}} \\
Nonclass           &     1127 &     29.9 & \\
Exchange 5k        &     1433 &     31.2 &     29.8 \\[1mm]
\multicolumn{4}{l}{\textbf{Finnish, 133k subwords}} \\
Nonclass           &     1135 &     30.1 & \\
Exchange 5k        &     1412 &     31.4 &     30.2 \\[1mm]
\multicolumn{4}{l}{\textbf{Finnish, 265k subwords}} \\
Nonclass           &     1128 &     30.2 & \\
Exchange 5k        &     1334 &     30.6 &     29.2 \\[1mm]
\multicolumn{4}{l}{\textbf{Finnish, 468k subwords}} \\
Nonclass           &     1100 &     30.0 & \\
Exchange 5k        &     1252 &     30.2 &     29.1 \\[1mm]
\hline
\noalign{\vskip 1mm}
\multicolumn{4}{l}{\textbf{Estonian, 1.8M words}} \\
Nonclass           &      447 &     23.4 & \\
Brown 2k           &      438 &     22.8 & \bf 22.5 \\
Exchange 2k        &      439 &     22.9 &     22.6 \\
Brown+Exchange 2k  &      438 & \bf 22.6 & \bf 22.5 \\
Brown 5k           &      432 &     22.7 & \bf 22.5 \\
Exchange 5k        &      432 &     23.0 &     22.6 \\
Brown+Exchange 5k  &  \bf 430 &     22.8 &     22.7 \\[1mm]
\multicolumn{4}{l}{\textbf{Estonian, 33.2k subwords}} \\
Nonclass           &      591 &     23.4 & \\
Exchange 5k        &      707 &     24.3 &     23.9 \\[1mm]
\multicolumn{4}{l}{\textbf{Estonian, 103k subwords}} \\
Nonclass           &      582 &     23.7 & \\
Exchange 5k        &      689 &     24.1 &     23.5 \\[1mm]
\multicolumn{4}{l}{\textbf{Estonian, 212k subwords}} \\
Nonclass           &      577 &     23.1 & \\
Exchange 5k        &      659 &     23.6 &     23.2 \\[1mm]
\multicolumn{4}{l}{\textbf{Estonian, 403k subwords}} \\
Nonclass           &      582 &     23.4 & \\
Exchange 5k        &      644 &     23.4 &     23.0 \\[1mm]
\hline
\end{tabular}
\end{center}
\end{table}

Table \ref{tab:ngram-dev-results} lists perplexities and word error rates given
by n-gram language models on the development data. The baseline word model
performance, 30.5 \% on Finnish and 23,4 \% on Estonian can be compared to the
various word class and subword models. We have also included results from
subword classes created using the exchange algorithm for reference. The word
error rates were obtained by rescoring lattices that were created using the
nonclass word or subword model.

In the Finnish task, 2,000, 5,000, and 10,000 class vocabularies were compared.
The worst case running times of the exchange and Brown algorithms have quadratic
dependency on the number of classes. With 10,000 classes, even using 20 CPUs
Brown did not finish in 20 days, so the experiment was not continued. The
exchange algorithm can be stopped any time, so there is no upper limit on the
number of classes that can be trained, but the quality of the clustering may
suffer if it is stopped early. However, with 5,000 classes and using only 5
CPUs, it seemed to converge in 5 days. Increasing the number of threads
increases memory consumption. Training even 40,000 classes was possible in 15
days, but the results did not improve, so they are not reported. The most
promising models were evaluated also in the Estonian task.

CBOW is clearly the fastest algorithm, which is probably why it has gained some
popularity. These results show, however, that the clusters formed by k-means
from distributed word representations are not good for n-gram language models.
CBOW does improve, compared to the other clusterings, when the number of classes
is increased. Other than that, the differences between different classification
methods are mostly insignificant, but class-based models outperform word models
on both languages. This result suggests that class-based softmax may be a viable
alternative to other softmax approximations in neural networks. The performance
of the Estonian subword models is close to that of the word model, and the
Finnish subword models are better than the word model. Subword classes do not
work as well, but the difference to nonclass subword models gets smaller
when the size of the subword vocabulary increases.

Mostly the differences between the different initializations of the exchange
algorithm seemed insignificant. However, our rule-based clustering algorithm
followed by running the exchange algorithm to create 2,000 classes
(Rules+Exchange 2k) gave the best word error rate on Finnish. In the NNLM
experiments we did not explore with different clusterings, but used the ones
that gave the smallest development set perplexity in the n-gram experiments. For
Finnish the 5,000 classes created using the exchange algorithm was selected
(Exchange 5k). On Estonian, initialization using Brown classes gave slightly
better perplexity (Brown+Exchange 5k) and was selected for neural network
models.

\begin{table}[t]
\caption{{\it Comparison of uniform data processing, random sampling of web
data by 20~\%, and weighted parameter updates from web data by a factor of 0.4,
in NNLM training. The models were trained using normal softmax. Includes
development set perplexity, word error rate (\%), and word error rate after
interpolation with the n-gram model.}}
\label{tab:subset-processing-results}
\begin{center}
\begin{tabular}{lcccc}
\hline
Subset     & Training \\
Processing & Time     & Perplexity &   WER & +NGram \\
\hline
\noalign{\vskip 1mm}
\multicolumn{5}{l}{\textbf{Finnish, 5k classes}} \\
Uniform    &    143 h &     511 &     26.0 &     25.6 \\
Sampling   &    128 h &     505 &     26.2 &     25.6 \\
Weighting  &    101 h &     521 &     26.4 &     25.5 \\[1mm]
\multicolumn{5}{l}{\textbf{Finnish, 42.5k subwords}} \\
Uniform    &    360 h &     679 &     25.2 & \bf 24.6 \\
Sampling   &    360 h &     671 &     25.5 &     25.0 \\
Weighting  &    360 h &     672 & \bf 25.1 & \bf 24.6 \\[1mm]
\multicolumn{5}{l}{\textbf{Finnish, 468k subwords, 5k classes}} \\
Uniform    &    141 h &     790 &     26.0 &     25.0 \\
Sampling   &    119 h &     761 &     25.9 &     25.1 \\[1mm]
\hline
\noalign{\vskip 1mm}
\multicolumn{5}{l}{\textbf{Estonian, 5k classes}} \\
Uniform    &     86 h &     339 & \bf 19.8 &     19.9 \\  
Sampling   &     87 h &     311 &     20.2 &     19.9 \\
Weighting  &    105 h &     335 &     20.0 & \bf 19.6 \\[1mm]
\multicolumn{5}{l}{\textbf{Estonian, 212k subwords, 5k classes}} \\
Uniform    &    187 h &     424 &     20.0 &     19.7 \\
Sampling   &    130 h &     397 &     20.0 &     19.8 \\
Weighting  &    187 h &     409 &     19.9 & \bf 19.6 \\[1mm]
\hline
\end{tabular}
\end{center}
\end{table}

Table \ref{tab:subset-processing-results} compares training time, perplexity,
and word error rate in NNLM training, when different processing is applied to
the large web data set. \textit{Uniform} means that the web data is processed
just like other data sets, \textit{sampling} means that a subset of web data is
randomly sampled before each epoch, and \textit{weighting} means that the
parameter updates are given a smaller weight when the mini-batch contains web
sentences. Sampling seems to improve perplexity, but not word error rate.
Because sampling usually speeds up training considerably and our computational
resources were limited, the rest of the experiments were done using sampling.

\begin{table}[t]
\caption{{\it NNLM results on development data. Word models were trained using
class and shortlist vocabularies. Subword models were trained using the full
subword vocabulary and on classes created from the subwords. Includes
perplexity, word error rate (\%), and word error rate after interpolation with
the n-gram model. Perplexities from word and subword models are not
comparable.}}
\label{tab:nn-dev-results}
\begin{center}
\begin{tabular}{l@{\hskip 0.3cm}l@{\hskip 0.3cm}c@{\hskip 0.3cm}c@{\hskip 0.3cm}
                c@{\hskip 0.3cm}c@{\hskip 0.1cm}c}
\hline
Network &             &            & Training \\
Output  & Vocabulary  & Parameters & Time & PPL &      WER & +NGram \\
\hline
\noalign{\vskip 1mm}
\multicolumn{7}{l}{\textbf{Finnish, 2.4M words}} \\
HSoftmax & 100k short & 231M &  360 h &     535 &     26.9 &     25.9 \\
NCE      & 100k short & 230M &  360 h &     531 &     26.8 &     25.7 \\
HSoftmax & 500k short & 532M &  360 h &     686 &     28.4 &     27.0 \\
Softmax  & 5k classes &  40M &  128 h &     505 &     26.2 &     25.6 \\[1mm]
\multicolumn{7}{l}{\textbf{Finnish, 42.5k subwords}} \\
Softmax  & 42.5k full & 115M &  360 h &     671 & \bf 25.5 & \bf 25.0 \\
HSoftmax & 42.5k full & 116M &  360 h &     700 &     25.7 & \bf 25.0 \\
Softmax  & 5k classes &  40M &  146 h &     857 &     26.2 &     25.5 \\[1mm]
\multicolumn{7}{l}{\textbf{Finnish, 133k subwords}} \\
HSoftmax & 133k full  & 164M &  360 h &     742 &     26.5 &     25.4 \\
Softmax  & 5k classes &  40M &  190 h &     811 &     26.1 &     25.4 \\[1mm]
\multicolumn{7}{l}{\textbf{Finnish, 265k subwords}} \\
HSoftmax & 265k full  & 296M &  360 h &     849 &     27.0 &     25.6 \\
Softmax  & 5k classes &  40M &  133 h &     813 &     26.2 &     25.3 \\[1mm]
\multicolumn{7}{l}{\textbf{Finnish, 468k subwords}} \\
HSoftmax & 468k full  & 500M &  360 h &    1026 &     28.6 &     26.9 \\
Softmax  & 5k classes &  40M &  119 h &     761 &     25.9 &     25.1 \\[1mm]
\hline
\noalign{\vskip 1mm}
\multicolumn{7}{l}{\textbf{Estonian, 1.8M words}} \\
HSoftmax & 100k short & 231M &  360 h &     321 &     20.6 &     19.9 \\
NCE      & 100k short & 230M &  142 h &     384 &     22.4 &     21.4 \\
HSoftmax & 500k short & 532M &  360 h &     380 &     21.0 &     20.2 \\
Softmax  & 5k classes &  40M &   87 h &     311 &     20.2 &     19.9 \\[1mm]
\multicolumn{7}{l}{\textbf{Estonian, 33.2k subwords}} \\
Softmax  & 33.2k full &  97M &  360 h &     357 &     20.4 &     20.2 \\
HSoftmax & 33.2k full &  97M &  293 h &     370 &     20.7 &     20.2 \\
Softmax  & 5k classes &  40M &  116 h &     418 &     20.9 &     20.2 \\[1mm]
\multicolumn{7}{l}{\textbf{Estonian, 103k subwords}} \\
HSoftmax & 103k full  & 134M &  306 h &     393 &     20.8 &     20.2 \\
Softmax  & 5k classes &  40M &  126 h &     410 &     20.5 &     19.9 \\[1mm]
\multicolumn{7}{l}{\textbf{Estonian, 212k subwords}} \\
HSoftmax & 212k full  & 243M &  360 h &     411 &     20.9 &     20.2 \\
Softmax  & 5k classes &  40M &  130 h &     397 & \bf 20.0 &     19.8 \\[1mm]
\multicolumn{7}{l}{\textbf{Estonian, 403k subwords}} \\
HSoftmax & 403k full  & 434M &  360 h &     463 &     21.4 &     20.7 \\
Softmax  & 5k classes &  40M &  124 h &     395 &     20.3 & \bf 19.6 \\[1mm]
\hline
\end{tabular}
\end{center}
\end{table}

Table \ref{tab:nn-dev-results} lists perplexities and word error rates
given by neural network models on the development data. The word error rates
were obtained by rescoring the same lattices as in Table
\ref{tab:ngram-dev-results}. The shortlist and word class models can predict all
training set words, so the perplexities can be compared. Subword models can
predict also new words, so their perplexities cannot be compared with word
models. The percentage of evaluation set words that are not in the shortlist and
words that are not in the training set can be found in Table
\ref{tab:oov-rates}.

The class-based models were clearly the fastest to converge, 5 to 8 days on
Finnish data and 4 to 6 days on Estonian data. The experiments include
shortlists of 100k and 500k words. Other shortlist models, except the Estonian
100k-word NCE, did not finish before the 360 hour limit. Consequently,
improvement was not seen from using a larger 500k-word shortlist.

Our NCE implementation required more GPU memory than hierarchical softmax and we
were unable to run it with the larger shortlist. With the smaller shortlist NCE
was better on Finnish and hierarchical softmax was better on Estonian. We
experienced issues with numerical stability using NCE with subwords, and decided
to use only hierarchical softmax in the subword experiments. BlackOut training
was slightly faster than NCE, but even less stable, and we were unable to finish
the training without numerical errors. With hierarchical softmax we used the
same classes that were used in the class-based models, but the classes were
rearranged to have equal sizes as described in Section
\ref{sec:hierarchical-softmax}. This kind of class arrangement did not seem to
improve from simple frequency binning, however.

In terms of word error rate and perplexity, class-based word models performed
somewhat better than the shortlist models. The best results were from subword
models. On both languages it can be seen that class-based subword models improve
compared to the nonclass subword models when the vocabulary size grows. In the
Finnish task, the smallest 42.5k-subword vocabulary worked well, which is small
enough to use normal softmax without classes. In the Estonian task, larger
subword vocabularies performed better, provided that the subwords were clustered
into classes. The best result was obtained by clustering 403k subwords into
5,000 classes using the exchange algorithm.

\begin{table}[t]
\caption{{\it Performance of best n-gram and NNLM models on evaluation data. All
NNLMs except the shortlist models were using softmax output. Includes
perplexity, word error rate (\%), and word error rate after interpolation with
the nonclass or n-gram model. Perplexities from word and subword models are not
comparable.}}
\label{tab:results-summary}
\begin{center}
\begin{tabular}{lcc@{\hskip 0.1cm}ccc@{\hskip 0.0cm}c}
\hline
                   & \multicolumn{3}{@{}c@{}}{N-Gram}    & \multicolumn{3}{@{}c}{NNLM} \\
Vocabulary         &          PPL &     WER & +Nonclass  &     PPL &     WER &   +NGram \\
\hline
\noalign{\vskip 1mm}
\textbf{Finnish}   & \multicolumn{3}{@{}c@{}}{Full 2.4M} & \multicolumn{3}{@{}c}{Shortlist 100k (NCE)} \\
Word               &          785 &     31.7 &           &     618 &     28.1 &     27.9 \\[1mm]
           & \multicolumn{3}{@{}c@{}}{Rules+Exchange 2k} & \multicolumn{3}{@{}c}{Exchange 5k} \\
Class              &          760 & \bf 31.4 &  \bf 31.1 &     589 &     29.2 &     27.9 \\[1mm]
             & \multicolumn{3}{@{}c@{}}{Morfessor 42.5k} & \multicolumn{3}{@{}c}{Morfessor 42.5k} \\
Subword            &         1313 &     31.7 &           &     846 & \bf 27.3 & \bf 27.1 \\[1mm]
              & \multicolumn{3}{@{}c@{}}{Morfessor 468k} & \multicolumn{3}{@{}c}{Morfessor 468k} \\
Subword Class      &         1499 &     32.1 &      31.3 &     942 &     28.2 &     27.4 \\[1mm]
\hline
\noalign{\vskip 1mm}
\textbf{Estonian}  & \multicolumn{3}{@{}c@{}}{Full 1.8M} & \multicolumn{3}{@{}c}{Shortlist 100k (HSoftmax)} \\
Word               &          483 &     26.1 &           &     344 &     23.1 &     22.6 \\[1mm]
           & \multicolumn{3}{@{}c@{}}{Brown+Exchange 2k} & \multicolumn{3}{@{}c}{Brown+Exchange 5k} \\
Class              &          465 & \bf 25.3 &  \bf 25.2 &     324 &     22.2 &     22.2 \\[1mm]
              & \multicolumn{3}{@{}c@{}}{Morfessor 212k} & \multicolumn{3}{@{}c}{Morfessor 33.2k} \\
Subword            &          628 &     26.0 &           &     377 &     22.7 &     22.6 \\[1mm]
              & \multicolumn{3}{@{}c@{}}{Morfessor 403k} & \multicolumn{3}{@{}c}{Morfessor 403k} \\
Subword Class      &          682 &     25.8 &      25.5 &     403 & \bf 22.1 & \bf 21.9 \\[1mm]
\hline
\end{tabular}
\end{center}
\end{table}

Table \ref{tab:results-summary} compares the best models on evaluation data. The
best word, class, subword, and subword class n-gram and NNLM models were
selected based on development set word error rate. The evaluation set results
show that the advantage that the Finnish subword n-gram models had on the
development set was due to the optimization of the morph segmentations on
development data. Word classes are the best choice for n-gram modeling. However,
neural networks seem to benefit especially subword modeling, because the overall
best results are from subword NNLMs. Classes work well also in NNLMs, although
the best Finnish shortlist model, 100k-word NCE, performed exceptionally well in
speech recognition. Interpolation with the n-gram model gives a small but
consistent improvement.

\section{Conclusions}

Our experiments show that class-based models are very attractive for
conversational Finnish and Estonian speech recognition. When the vocabulary
contains millions of words, class-based n-gram models perform better than normal
word models. A class-based NNLM can be trained in less than a week and when the
training time is limited, often performs better than word-shortlist models.

In previous work, class-based models did not outperform word-based models in
recognizing standard Finnish and Estonian, such as broadcast news
\cite{Varjokallio:2016}. Improvement was made only when interpolating word and
class models. One reason why word classes are especially beneficial in the
conversational tasks may be that in the absence of large conversational corpora,
most of our training data is from the Internet. Web data is noisy and there are
many ways to write the same word.

One would expect less to be gained from using subword models, when a word model
is trained from full vocabulary of millions of words. This seems to be the case,
but RNNs are good at learning the structure of the language from a text that has
been segmented into subwords. Subwords can also solve the vocabulary size
problem with neural network models. In the Finnish task, the best results were
from an NNLM trained on a relatively small 42.5k-subword vocabulary with full
softmax output. In the Estonian task, the best results are from a large
403k-subword vocabulary that was clustered into 5,000 classes.

We explored the possibility of using NCE, BlackOut, or hierarchical softmax to
overcome the problem of training neural networks with large output
dimensionality. Generally they were slower than class-based training, and did
not converge to as good a model in the 15-day time constraint, but Finnish
100k-word NCE training gave good results on the evaluation set. The mixed
results could mean that some details have been overlooked in our implementation
of sampling-based softmax.

In both tasks we obtained the best word error rates from a subword NNLM
interpolated with a subword n-gram model. In the Finnish task the best result
was 27.1 \%, which is a 14.5 \% relative improvement from the 31.7 \% WER given
by our baseline 4-gram model. The best result in the Estonian task, 21.9 \%, is
a 16.1 \% relative improvement from our 26.1 \% baseline WER. These are the best
results achieved in these tasks, and better than our previously best results by
a large margin. The best previously published results are 48.4 \% WER in the
Finnish task \cite{Enarvi:2016} and 52.7 \% WER in the Estonian task
\cite{Kurimo:2016}.

The corpus weighting methods that we used in NNLM training showed potential for
improvement, but more thorough research should be done on how to select optimal
weights.

\section{Acknowledgements}

Computational resources were provided by the Aalto Science-IT project.

\bibliographystyle{IEEEtran}
\bibliography{IEEEabrv,references}

\begin{IEEEbiography}
[{\includegraphics[width=1in,height=1.25in,clip,keepaspectratio]{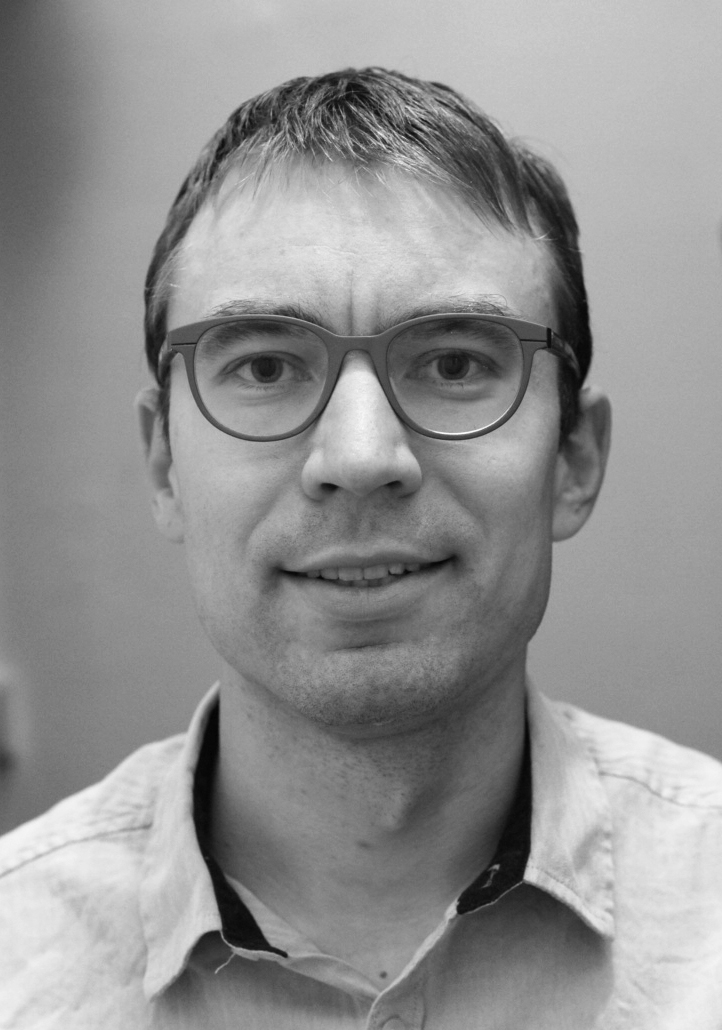}}]
{Seppo Enarvi} received the Lic.Sc. in technology degree in computer and
information science from Aalto University, Espoo, Finland, in 2012. He pursues
to finish his Ph.D. on conversational Finnish speech recognition during 2017.
His research interests are in machine learning, currently focusing on language
modeling using neural networks.
\end{IEEEbiography}

\begin{IEEEbiography}
[{\includegraphics[width=1in,height=1.25in,clip,keepaspectratio]{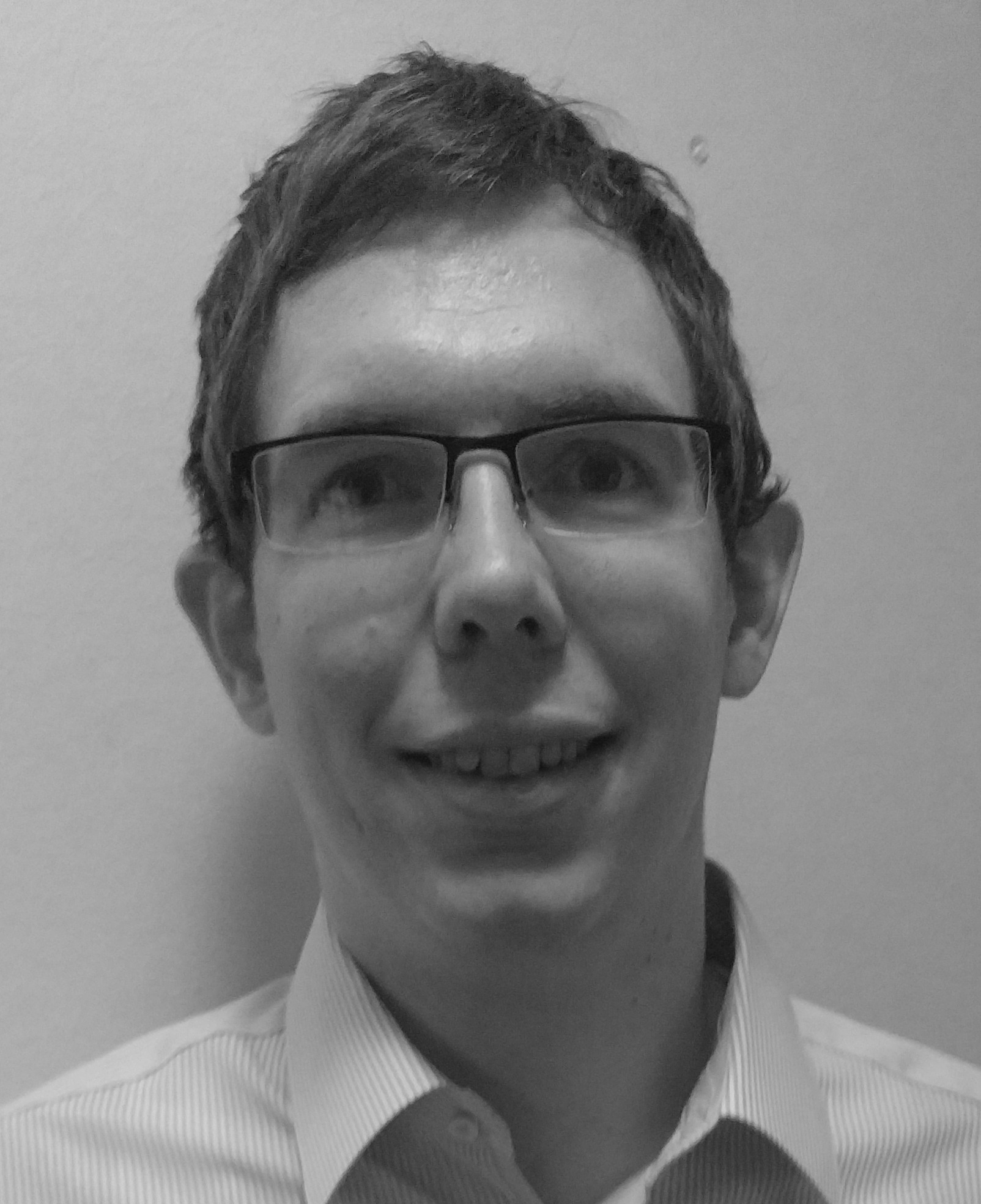}}]
{Peter Smit} received the M.Sc. in technology degree in computer science from
Aalto University, Espoo, Finland, in 2011. He is currently a doctoral student in
the Department of Signal Processing and Acoustics at Aalto University. His
research interests are in machine learning and speech recognition and his
current focus is subword-modeling techniques and automatic speech recognition
for underresourced languages.
\end{IEEEbiography}

\begin{IEEEbiography}
[{\includegraphics[width=1in,height=1.25in,clip,keepaspectratio]{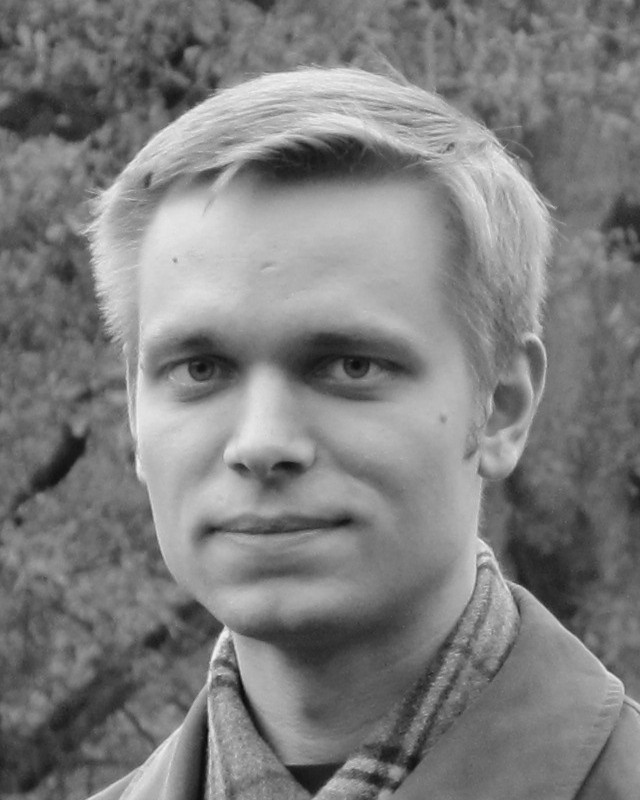}}]
{Sami Virpioja} received the D.Sc. in technology degree in computer and
information science from Aalto University, Espoo, Finland, in 2012. Between
2013 and 2017 was a research scientist at Lingsoft Inc., Helsinki, Finland.
Currently he is a senior data scientist at Utopia Analytics Oy, Helsinki,
Finland, and a postdoctoral researcher in the Department of Signal
Processing and Acoustics at Aalto University. His research interests
are in machine learning and natural language processing.
\end{IEEEbiography}

\begin{IEEEbiography}
[{\includegraphics[width=1in,height=1.25in,clip,keepaspectratio]{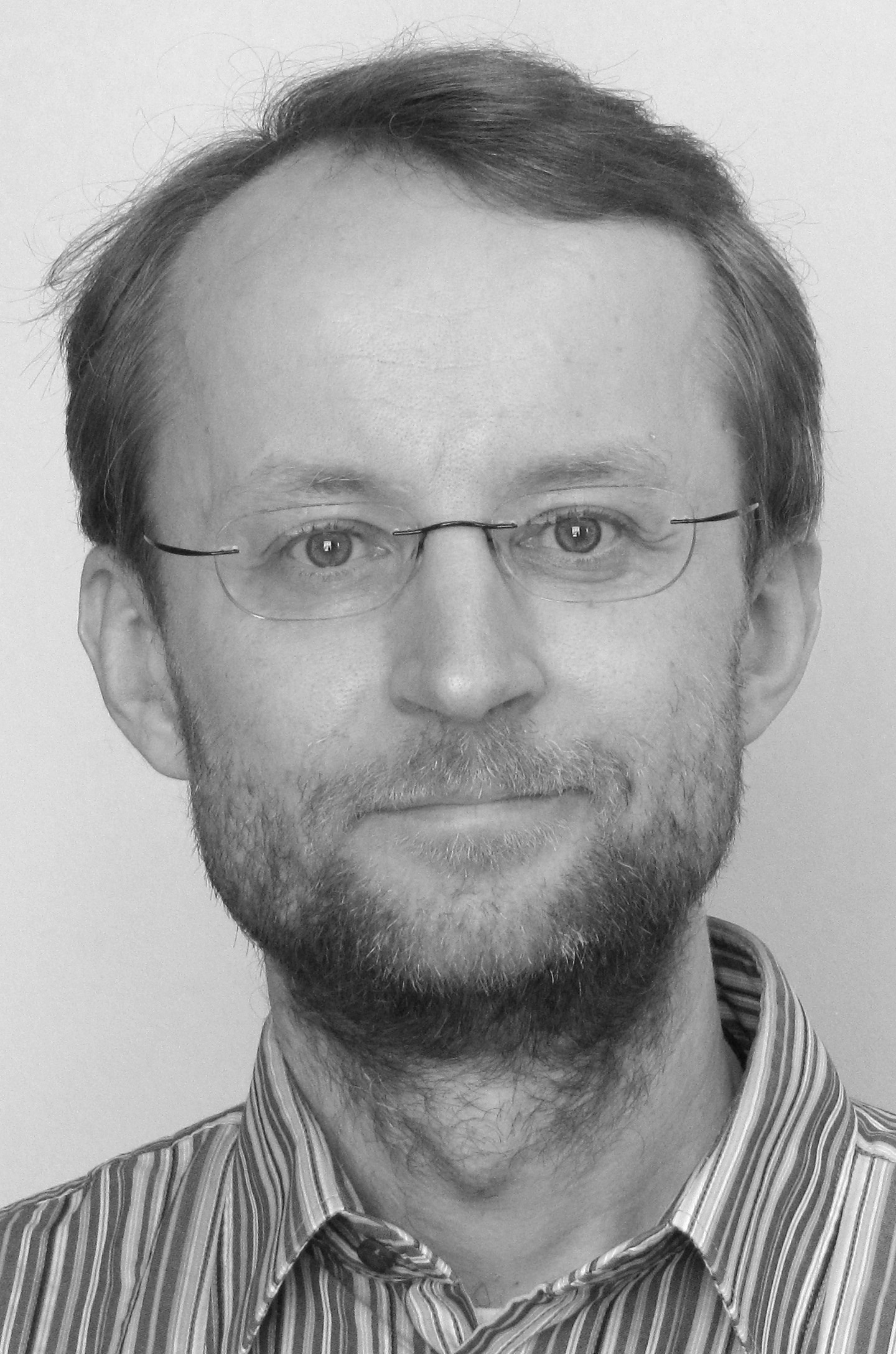}}]
{Mikko Kurimo} (SM'07) received the D.Sc. (Ph.D.) in technology degree in
computer science from the Helsinki University of Technology, Espoo, Finland, in
1997. He is currently an associate professor in the Department of Signal
Processing and Acoustics at Aalto University, Finland. His research interests
are in speech recognition, machine learning and natural language processing.
\end{IEEEbiography}

\pagebreak

\hrulefill
\section*{Erratum}

A highway network layer uses a separate bias for its gate, distinguished by the
index $\sigma$ in Equation~\ref{eq:highway-network}. The index was missing in
the published paper.

\end{document}